\documentclass{article}
\usepackage{arxiv}
\usepackage[utf8]{inputenc} 
\usepackage[T1]{fontenc}    
\usepackage{hyperref}       
\usepackage{url}            
\usepackage{booktabs}       
\usepackage{amsfonts}       
\usepackage{nicefrac}       
\usepackage{microtype}      
\usepackage{lipsum}
\usepackage{amsmath}
\usepackage{amssymb}
\usepackage{graphicx}
\usepackage{textcomp}

\DeclareMathOperator{\Var}{Var}
\DeclareMathOperator{\E}{E}
\DeclareMathOperator{\Bias}{Bias}

\title{Physics Enhanced Artificial Intelligence}

\author{Patrick~O'Driscoll \\
Rebellion Photonics\\
Houston, TX 77002 \\
\texttt{patricko@rebellionphotonics.com} \\
   \And
 Jaehoon~Lee\\
 Rebellion Photonics\\
 Houston, TX 77002 \\
 \texttt{jaehoon@rebellionphotonics.com} \\
   \AND
 Bo Fu \\
 Rebellion Photonics\\
 Houston, TX 77002 \\
 \texttt{bo@rebellionphotonics.com} \\
}

\begin{document}
 
\maketitle

\begin{abstract}
    We propose that intelligently combining models from the domains of Artificial Intelligence or Machine Learning with Physical and Expert models will yield a more ``trustworthy'' model than any one model from a single domain, given a complex and narrow enough problem. Based on mean-variance portfolio theory and bias-variance trade-off analysis, we prove combining models from various domains produces a model that has lower risk, increasing user trust. We call such combined models - physics enhanced artificial intelligence (PEAI), and suggest use cases for PEAI.
\end{abstract}


\section{ Introduction }

\subsection{ Motivation }

Recent theoretical developments in machine learning (ML), complemented by the astounding growth of computational power and the genesis of large data sets, have contributed to the rapid development of artificial intelligent (AI) systems. Even though the key findings required for a general AI system (strong AI) are considered a distant endeavor \cite{Muller2016}, AI systems designed to solve narrowly defined yet challenging enough problems (weak AI or narrow AI) are often comparable to or exceeding the performance of average humans \cite{Yu2017,Lu2015SurpassingHF}, and in many cases human experts, at these same tasks \cite{He_2015_ICCV,Silver2016,AssaelSWF16,Brown418}.
These narrow AI solutions offer a great potential for industry to automate, improve, and surpass unaided human productivity. In the rest of this paper, the term AI will refer to this narrow AI, unless stated otherwise. In spite of the great performance potential AI encompasses, user adoption has always been challenging. In many cases user trust becomes a bottleneck towards industry-wide adoption, especially in aerospace, safety, and defence, to name a few. New ways to enhance user trust in AI can directly affect user adoption at a 
large scale. In addition, if new ways to enhance user trust in AI can take advantage of existing solutions that were developed prior to AI solutions, it is likely to be more resource-friendly and even more attractive to industry. 

\subsection{ Background and Related Work }

In pursuit of industry-wide adoption of AI, new areas of research that focus on the trustworthiness of AI have emerged. Trust is a topic of rich content deeply rooted in many historical and philosophical discussions, and is often tied with the study of risk in philosophical research \cite{Roeser2012}. As we are primarily interested in user adoption of new technology, we are not approaching trust from a philosophical research viewpoint and focus on the aspect of user trust. To many ordinary users, the lack of trust in AI may have originated from the perception of the technology as a `black box'. This perception reflects several other profound issues between human users and AI, including lack of understanding of the scientific principles of how the AI is constructed, lack of understanding of the functionality and limitations of ML based systems, and lack of transparency in the AI design process. 
Even for experts, the lack of straightforward ways of explanation of AI action using domain knowledge can lead to a `black box' perception. Recently there are considerable research efforts on developing AI systems that are easily interpreted by humans, resulting in an emerging research field of e\underline{X}plainable \underline{A}rtificial \underline{I}ntelligence, or XAI  \cite{Adadi2018,Mohseni2018ASO, Dosilovic2018,Biran2017ExplanationAJ}. XAI aims to bridge the gap of trust between AI system and its users by providing explanation of AI systems with the intentions to justify, control, and improve AI actions. Since explanations are subjective to the human observer, this area has also expanded to include psychology, philosophy, and cognitive science of people \cite{Miller2017ExplanationIA}.

The long term solution to enhancing user trust in AI is to progress theoretical understanding of the underlying principles of machine learning and data science, similar to how we gain trust in systems that are designed based on physical principles. Being a relatively young field, the AI community is beginning to address some of the issues that may hinder user trust, such as lack of empirical rigor of some of the works conducted by the community \cite{sculley2018winner}. Although questions in the fundamental level of AI and ML research still remains, especially in deep learning \cite{Davide2016}, it should be noted that this alone does not provide a direct cause to diminish trustworthiness of AI. There are open and outstanding questions within our mainstream theories of physics, as well as in our own understanding of the human brain and its functionality, and yet we generally trust systems design via physical principles and judgments from biological systems that we do not fully understand. 

Other ways to improve user trust and adoption in AI include providing tractable performance improvement compared to non-AI systems in controlled data set and experiments. This is normally achieved by building metrics to quantify performance \cite{Jiang2018ToTO}. However, any quantitative comparison will likely tie the performance to the testing data sets with known data bias being an issue for AI systems \cite{Jiang2019}. Another practical way to restoring user trust for AI services is by providing supplementary documentation such as a supplier's declaration of conformity (SDoC) for the AI services provided \cite{hind2018increasing}. 

In the above-mentioned efforts to build trustworthy AI systems, one aspect is overlooked: the existing trust of users in systems and solutions constructed using physics-based principles and other domain expert knowledge. In this paper we propose a new perspective to restore trust of AI by leveraging user trust in physical principles and expert knowledge, wherever applicable. By intelligently combining systems based on Artificial Intelligence or Machine Learning with Physical and Expert systems, we show that the overall system can have improved user trust. As the source of the enhanced trust comes from physics, we term this class of AI with enhanced trust the \underline{P}hysics \underline{E}nhanced \underline{A}rtificial \underline{I}ntelligence (PEAI). In the rest of this paper, we will develop this idea by leveraging well known theories such as portfolio theory \cite{markowitz1952} and bias-variance trade-off analysis, and discuss the implications for design of complex systems. 

\section{ Physics Enhanced Artificial Intelligence (PEAI) }

\subsection{ Definitions and Assumptions }

In order to discuss the mathematical description of PEAI, we first discuss the basic concepts behind the idea of trust and risk in this context. In general, a user is more likely to trust and adopt a new technology - presented in the form of a model, when it is explainable and has good performance. We therefore assume that user trust, for any given individual, is composed of three properties of the model:
\begin{enumerate}
    \item Interpretability or Explainablity ($E$)
    \item Performance Accuracy ($A$)
    \item Performance Consistency ($C$)
\end{enumerate}

In many practical applications where the task at hand is complex, AI models learned from data have lower interpretability compared with physics-based or expert-based models, and their consistency can be unknown or poor depending on how they are trained and the training data provided. However, they tend to be more accurate. Physics-based models are often quite interpretable and consistent, but are often not as accurate as the AI derived models. Expert models, while often accurate, may not be as consistent or explainable as physics-based models. We argue that combining models or knowledge from different domains of AI, physics, and expert for narrowly defined tasks, will yield a more trustworthy model than any one model or knowledge base they are composed from. Interpretability is subjective, and it is beyond the scope of our discussion of PEAI. By assuming that the interpretability of the models in question is constant based upon a given individual, we maximize user trust by maximizing the accuracy for a given model consistency. 

As an attempt to qualify the relationship between trust and a given model, we express user trust as 
\begin{equation}
    \text{User Trust} \propto UT(a,c). 
\end{equation}
where $a\in A$, $c\in C$, and $UT$ is a function of performance qualities that are assumed to belong to partially ordered domains.
Further, we assume that
\begin{equation}
    \sup\{ UT (a,c) \} = UT(\sup\{A\}, \sup\{C\}),
\end{equation}
and 
\begin{equation}
    \inf\{ UT (a,c) \} = UT(\inf\{A\}, \inf\{C\}),
\end{equation}
such that we can optimize $UT$.
PEAI aims to develop a model to solve a narrow problem which consists of a set of rules and requirements that is described by task $\mathcal{T}$. To avoid the discussion of trivial and unreasonable situations, we assume that $\mathcal{T}$ is sufficiently complex that the optimal solution is not known, and will require near infinite resources to identify. In order to solve $\mathcal{T}$, a model, or a solution, $f: \mathcal{X} \mapsto \mathcal{Y}_{\mathcal{T}}$ is constructed, where $\mathcal{X}$ consists of inputs from sensor measurements and $\mathcal{Y}_{\mathcal{T}}$ its outputs. Let $\mathcal{U}\neq{\emptyset}$ be the universal set of solutions. As models can be constructed using different methods, we define the following:
\begin{align} 
    \mathcal{A}   &= \{ f_\mathcal{A}\in \mathcal{U}  : f_\mathcal{A} \text{ constructed using only AI based methods } \} \\
    \mathcal{P}   &= \{ f_\mathcal{P}\in \mathcal{U}  : f_\mathcal{P} \text{ constructed using only physics based methods } \}\\
    \mathcal{E}   &= \{ f_\mathcal{E}\in \mathcal{U}  : f_\mathcal{E} \text{ constructed using only expert based methods } \}\\
    \mathcal{AP}  &= \{ f_\mathcal{AP}\in \mathcal{U} : f_\mathcal{AP} \text{ constructed using AI and physics based methods } \}\\
    \mathcal{AE}  &= \{ f_\mathcal{AE}\in \mathcal{U} : f_\mathcal{AE} \text{ constructed using AI and expert based methods } \}\\
    \mathcal{PE}  &= \{ f_\mathcal{PE}\in \mathcal{U} : f_\mathcal{PE} \text{ constructed using physics and expert based methods } \}\\
    \mathcal{APE} &= \{ f_\mathcal{APE}\in \mathcal{U}: f_\mathcal{APE} \text{ constructed using AI, physics, and expert methods } \}.
\end{align}

We also assume that there exist multiple competing models from each of the above defined sets. A PEAI system (or model), $f_{PEAI}$, belongs to the set
\begin{align}
         \mathcal{S} &= \mathcal{APE} \cup \mathcal{AP} \cup \mathcal{AE} \cup ( \mathcal{A}\cap\mathcal{AP} )\cup 
     (\mathcal{A}\cap\mathcal{AE} )  \cup ( \mathcal{A}\cap\mathcal{PE} )\cup (\mathcal{A}\cap\mathcal{P} ) \cup ( \mathcal{A}\cap\mathcal{E} )\\
 &= \mathcal{APE} \cup \mathcal{AP} \cup \mathcal{AE} \cup  ( \mathcal{A}\cap\mathcal{PE} )\cup (\mathcal{A}\cap\mathcal{P} ) \cup ( \mathcal{A}\cap\mathcal{E} ).
\end{align}

For a complex task $\mathcal{T}$, it is highly unlikely that one arrives at exactly the same mathematical model when using different modeling methods, therefore the intersections between any two sets of model sets are expected to be empty, i.e, $\mathcal{A}\cap\mathcal{P},\mathcal{A}\cap\mathcal{E},\mathcal{A}\cap\mathcal{PE}=\emptyset$. Under these situations, $\mathcal{S}$ reduces to
\begin{equation} \label{eq:f_PEAI}
   \mathcal{S}^* =\mathcal{APE} \cup \mathcal{AP} \cup \mathcal{AE}.
\end{equation}
We will examine $f_{PEAI} \in \mathcal{S}^*$ for the remainder of this paper and discuss strategies for constructing $f_{PEAI}$. 
For a complex problem, models are expensive to make, and no one model is perfect. We consider composing $N$ finite number of models. Finally we assume that all models that are examined in $\mathcal{U}$ are constructed in good faith and aim to provide the best results possible given their application and method.

\subsection{Construction of PEAI}

There are two strategies to make a PEAI algorithm. The first is a composite model output approach - take models from the sets $\mathcal{A}$, $\mathcal{P}$, and $\mathcal{E}$, and combine their outputs to form a new composite model in $\mathcal{S^*}$. The composite model approach will be analyzed using an analogy to classical mean-variance portfolio theory. The second is a hybridization model approach - modify the form of constructing the model by applying an intelligent constraint using information from another domain, generating a model in $\mathcal{S^*}$. We will analyze this hybridized model by using classical bias-variance trade-off analysis. We show that composed models using the above strategies yield a more consistent model for a desired accuracy. 

\subsubsection{ Composite PEAI using Mean-Variance Portfolio Theory }

The 1990 Nobel prize was awarded to Harry Markowitz for his 1952 `Portfolio Selection' essay \cite{markowitz1952}. His work laid the mathematical foundation of diversification by demonstrating that the combination of risky assets is less risky than any single asset. By treating the available models in $\mathcal{U}$ as risky assets, we can maximize user trust by minimizing the variance (risk) of the composite model. While this is conceptually a simple idea, it has profound impact on the understanding of ML ensembles and composite model techniques. 

Assume there exists a function $m$: $\mathcal{Y_\mathcal{T}} \mapsto \mathbb{R}^+$ that can be evaluated on each of the models that gives a meaningful representation of the model performance. 
Each model $i \in \{1, 2, ... , N \}$, has the output $Y_{i}$.
Without a loss of generality, we further assume larger value of $m(Y)$ indicate better performance.
For the combined model, we have
\begin{equation}
    m(Y_{C}) = \sum_i { w_i m(Y_i)},
\end{equation}
where $w_i$ is the relative weight given to model $i$ with $\sum_i w_i = 1$, and $w_i \geq 0,\,\, \forall i$. Therefore $C$ is composed of all $N$ models. Using the distributive property of the expectation denoted by $\E[\cdot]$, we solve for the first moment of $Y_C$, 
\begin{equation}
    \mu_C = \E \left[ m(Y_C) \right] = \sum_i \E[m(Y_i)] w_i = \boldsymbol{\mu}^T \mathbf{w},
\end{equation}
where $\mu_k = \E [ m(Y_k)]$, $\mathbf{w} = \left[ w_1\,\, w_2\,\, ...\, w_N \right]$, and $\boldsymbol{\mu} = \left[ \mu_1\,\, \mu_2\,\, ... \, \mu_N \right]$. We will assume that each element of $\boldsymbol{\mu}$ is not equal to each-other, as the models are expected to give different expected values.

The second moment of $m(Y_C)$,
\begin{equation} \label{eq:mva_sp}
    \sigma_C^2 = \Var \left[ m(Y_C) \right] = \mathbf{w}^T \boldsymbol{\Omega} \mathbf{w},
\end{equation}
where $\boldsymbol{\Omega}$ is the covariance matrix. We will assume that $\boldsymbol{\Omega}$ exists, as a result of the models each being unique rather than a composition of other models.

By minimizing the variance of the combined model, we are reducing risk. As the combined model is more likely to give a consistent answer. The minimization is subject to the following previously defined constraints: $\mu_C = \boldsymbol{\mu}^T \mathbf{w}$ and $\mathbf{1}^T \mathbf{w} = 1$, where $\mathbf{1} = [1\,\, 1\,\, ...\, 1]$.

To find the optimized weights, we construct the Largrangian, $L$:
\begin{equation}
    L = \mathbf{w}^T \boldsymbol{\Omega} \mathbf{w} - \lambda_1 \left( \mathbf{1}^T \mathbf{w} - 1 \right) - \lambda_2 \left( \boldsymbol{\mu}^T \mathbf{w} - \mu_C \right).
\end{equation}

Taking the partial derivatives of $L$ with respect to $w_i$, $\lambda_1$, and $\lambda_2$:
\begin{equation} \label{eq:mva_p_w}
    \frac{\partial L}{\partial \mathbf{w}} = \mathbf{w}^T \boldsymbol{\Omega} - \lambda_1 \mathbf{1}^T - \lambda_2 \boldsymbol{\mu}^T = \mathbf{0}
\end{equation}
\begin{equation} \label{eq:mva_p_l1}
    \frac{\partial L}{\partial \lambda_1} = \mathbf{1}^T \mathbf{w} - 1 = 0
\end{equation}
\begin{equation} \label{eq:mva_p_l2}
    \frac{\partial L}{\partial \lambda_2} = \boldsymbol{\mu}^T \mathbf{w} - \mu_C = 0
\end{equation}

By rearranging equation \ref{eq:mva_p_w},
\begin{equation}
    \mathbf{w}^* = \boldsymbol{\Omega}^{-1}(\lambda_1 \mathbf{1} + \lambda_2 \boldsymbol{\mu}),
\end{equation}
where $\mathbf{w}^*$ is the optimized weights that minimize $\sigma_C^2$.
From equations \ref{eq:mva_p_l1} and \ref{eq:mva_p_l2}:
\begin{equation} \label{eq:mva_lin_1}
    1 = \mathbf{1}^T \mathbf{w}^* = \mathbf{1}^T \boldsymbol{\Omega}^{-1} \boldsymbol{\Omega} \mathbf{w^*} = \lambda_1 \mathbf{1}^T \boldsymbol{\Omega}^{-1} \mathbf{1} + \lambda_2 \mathbf{1}^T \boldsymbol{\Omega}^{-1} \boldsymbol{\mu},
\end{equation}
\begin{equation} \label{eq:mva_lin_2}
    \mu_C = \boldsymbol{\mu}^T \mathbf{w}^* = \boldsymbol{\mu}^T \boldsymbol{\Omega}^{-1} \boldsymbol{\Omega} \mathbf{w}^* = \lambda_1 \boldsymbol{\mu}^T \boldsymbol{\Omega}^{-1} \mathbf{1} + \lambda_2 \boldsymbol{\mu}^T \boldsymbol{\Omega}^{-1} \boldsymbol{\mu}.
\end{equation}
Define new variables
\begin{equation} \label{eq:mva_abg}
    \begin{split}
        \alpha = \mathbf{1}^T \boldsymbol{\Omega}^{-1} \mathbf{1} \\
        \beta  = \mathbf{1}^T \boldsymbol{\Omega}^{-1} \boldsymbol{\mu} \\
        \gamma = \boldsymbol{\mu}^T \boldsymbol{\Omega}^{-1} \boldsymbol{\mu} \\
        \delta =  \alpha \gamma - \beta^2. \\
    \end{split}
\end{equation}

By substitution of the definitions from equation \ref{eq:mva_abg} into equations \ref{eq:mva_lin_1} and \ref{eq:mva_lin_2}. We can solve for $\lambda_1$ and $\lambda_2$ in terms of $\alpha$, $\beta$, and $\gamma$:
\begin{equation}
    \begin{split}
        \lambda_1 = \frac{\gamma - \beta \mu_C}{\delta},\\
        \lambda_2 = \frac{\alpha \mu_C - \beta}{\delta}.
    \end{split}
\end{equation}

Finally by substituting the above results into equation \ref{eq:mva_sp},
\begin{equation}\label{eq:v^2_c}
    \sigma_C^2 = \mathbf{w^*}^T \boldsymbol{\Omega} \mathbf{w^*} = \mathbf{w^*}^T \boldsymbol{\Omega} \left( \boldsymbol{\Omega}^{-1} \left( \lambda_1 \mathbf{1} + \lambda_2 \boldsymbol{\mu} \right) \right) = \lambda_1 + \lambda_2 \mu_C = \frac{\alpha \mu_C^2 - 2 \beta \mu_C + \gamma}{\delta}.
\end{equation}

It is important to note a few properties about $\alpha$, $\gamma$, and $\delta$. Since $\boldsymbol{\Omega}$ is positive definite, $\alpha > 0$ and $\gamma > 0$, and by the Cauchy-Schwarz inequality $\delta > 0$.

\begin{figure}[ht!]
    \centering
    \includegraphics[width=0.45\textwidth]{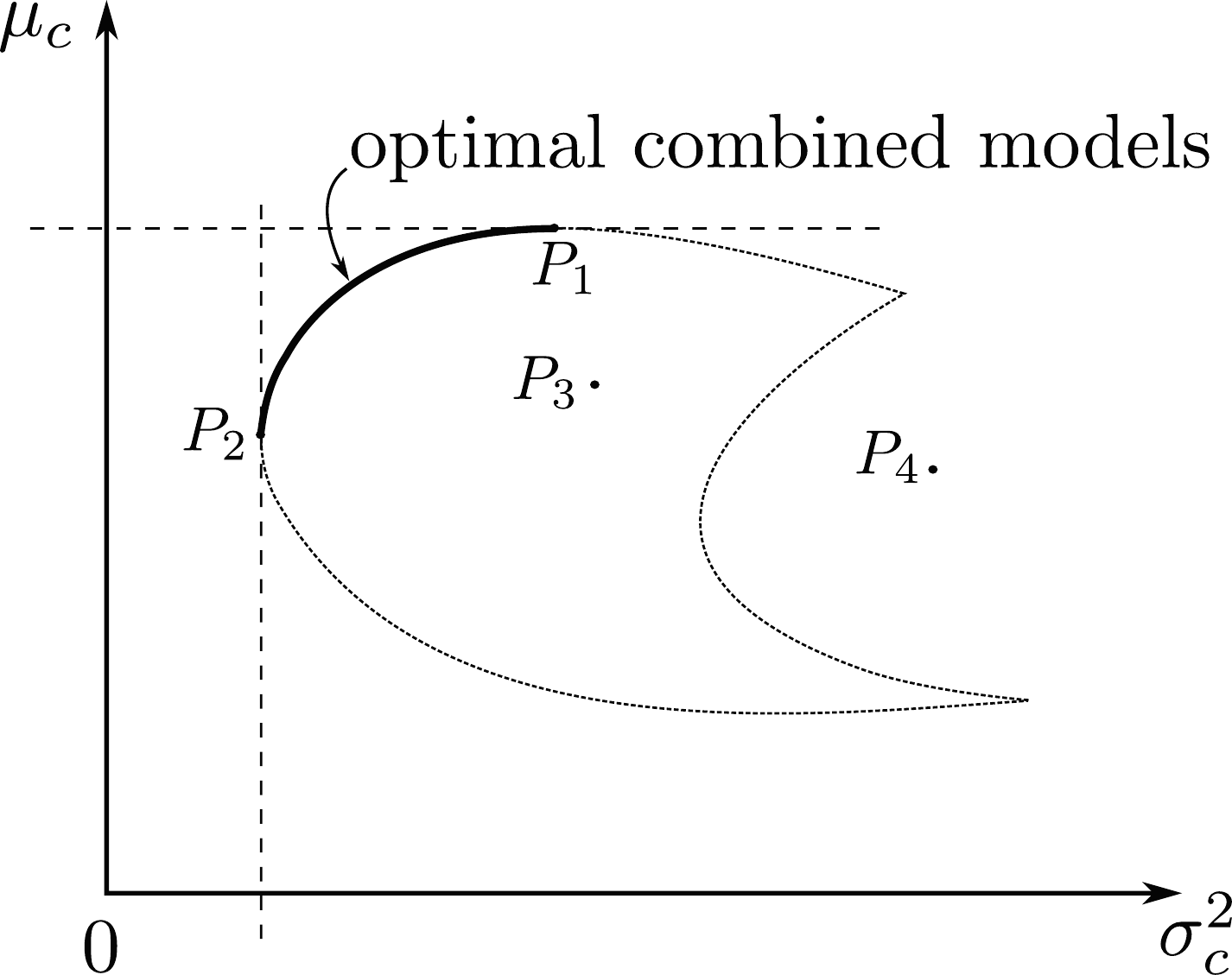}
    \caption{ Example of a feasible region of composite models from portfolio analysis. The optimal combined models available exist on the curve between points $P_1$ and $P_2$ and is marked by the thicker line. }
    \label{fig:p_svm}
\end{figure}

Equation \ref{eq:v^2_c} allows us to solve for the minimal risk for a desired mean value. For each $\mu_i$ being unique, the models not being perfectly correlated, and $N \geq 3$, the feasible region for portfolio theory can be shown to be a two-dimensional surface that is convex to the left, and is represented on the $\sigma_C^2$ vs. $\mu_C$ plane, see Figure \ref{fig:p_svm}. In this figure, the class of optimal combined models lines on the thick black line between points $P_1$ and $P_2$. A sub-optimal model found by combining model outputs from various domains, as shown by $P_3$, lies within the region with a dotted line boundary, and an unfeasible model lies outside this boundary, as shown by point $P_4$.

An example of the feasible point $P_3$ could be constructed and interpreted under the following conditions: If one were to make a model using a linear combination of the outputs of all $N$ models, and assign each model the weights of $\frac{1}{N}$, then one would construct the equivalent of an ensemble of models using majority vote to make a decision. In practice this has shown to increase the accuracy and reduce the variance of the model \cite{Chollet}; however, here we show that there exists a set of weights for each of these models that would minimize the risk of the prediction. Therefore, this ensemble, majority vote model is likely to be a sub-optimal solution for a given performance. 

In the case where models come from different domains, it is more likely that the models are going to have different $\mu_i$, be less correlated with each other, and have different properties of their predictions. For example, the models in $\mathcal{A}$ are more likely to be more accurate than those in $\mathcal{P}$, but have a higher variance. Therefore when these models are combined the composite model is able to obtain a more optimal performance than any one given model.

\subsubsection{ Hybridization PEAI using Bias-Variance Trade-Off Analysis }

Hybridized PEAI models can be shown to have lower risk and enhanced user trust. First we derive the expressions of bias and variance in a model. Here we will assume that the model can be represented as a function of the inputs plus some error:
\begin{equation}
    y = f(x) + \varepsilon.
\end{equation}
where $\varepsilon$ is noise with zero mean and variance $\sigma^2$, such that $\E[\varepsilon] = 0$ and $\E[\varepsilon^2] = \Var[y] = \sigma^2$. Also let $\hat{f}(x)$ be a deterministic approximation the function $f(x)$, $F = f(x)$ and $\hat{F} = \hat{f}(x)$. 

Therefore we can express the mean squared error as a function of the $Bias[\hat{F}]$, $\sigma^2$, and $\Var[\hat{F}]$:
\begin{equation}
    \begin{split}
        \E \left[ \left(y - \hat{F}\right)^2 \right] 
        & = \E\left[ \left( F + \varepsilon - \hat{F} \right)^2 \right] \\
        & = \E\left[ \left( F + \varepsilon - \hat{F} + \E[ \hat{F} ] - \E[ \hat{F} ] \right)^2 \right] \\
        & = \E\left[ \left( F - \E[\hat{F}] \right)^2 \right] + \E[\varepsilon^2] + 
            \E\left[ \left( \E[\hat{F}] - \hat{F} \right)^2 \right] + 
            2 \E\left[ \left( F - \E[\hat{F}] \right) \varepsilon \right] \\
            & \hspace{10mm} + 2 \E\left[ \varepsilon \left( \E[\hat{F}] - \hat{F}  \right) \right] + 
            2 \E\left[ \left( \E[\hat{F}] - \hat{F} \right) \left( F - \E[\hat{F}] \right) \right] \\
        & = \left( F - \E[\hat{F}] \right)^2 + \E\left[ \varepsilon^2 \right] + 
            \E\left[ \left( \E[ \hat{F} ] - \hat{F} \right)^2 \right] + 
            2 \left( F - \E[\hat{F}] \right)\E[\varepsilon] \\
            & \hspace{10mm} + 2 \E[\varepsilon] \E\left[ \E[\hat{F}] - \hat{F} \right] +
            2 \E \left[ \E[\hat{F}] - \hat{F} \right] \left( F - \E[\hat{F}] \right) \\
        & = \left( F - \E[\hat{F}] \right)^2 + \E\left[ \varepsilon^2 \right] + 
            \E\left[ \left( \E[\hat{F}] - \hat{F} \right)^2 \right] \\
        & = \left( F - \E[\hat{F}] \right)^2 + \sigma^2 + \Var[\hat{F}] \\
        & = \Bias[\hat{F}]^2 + \sigma^2 + \Var[\hat{F}]
    \end{split}
\end{equation}

\begin{figure}[ht!]
\centering
\includegraphics[width=0.5\textwidth]{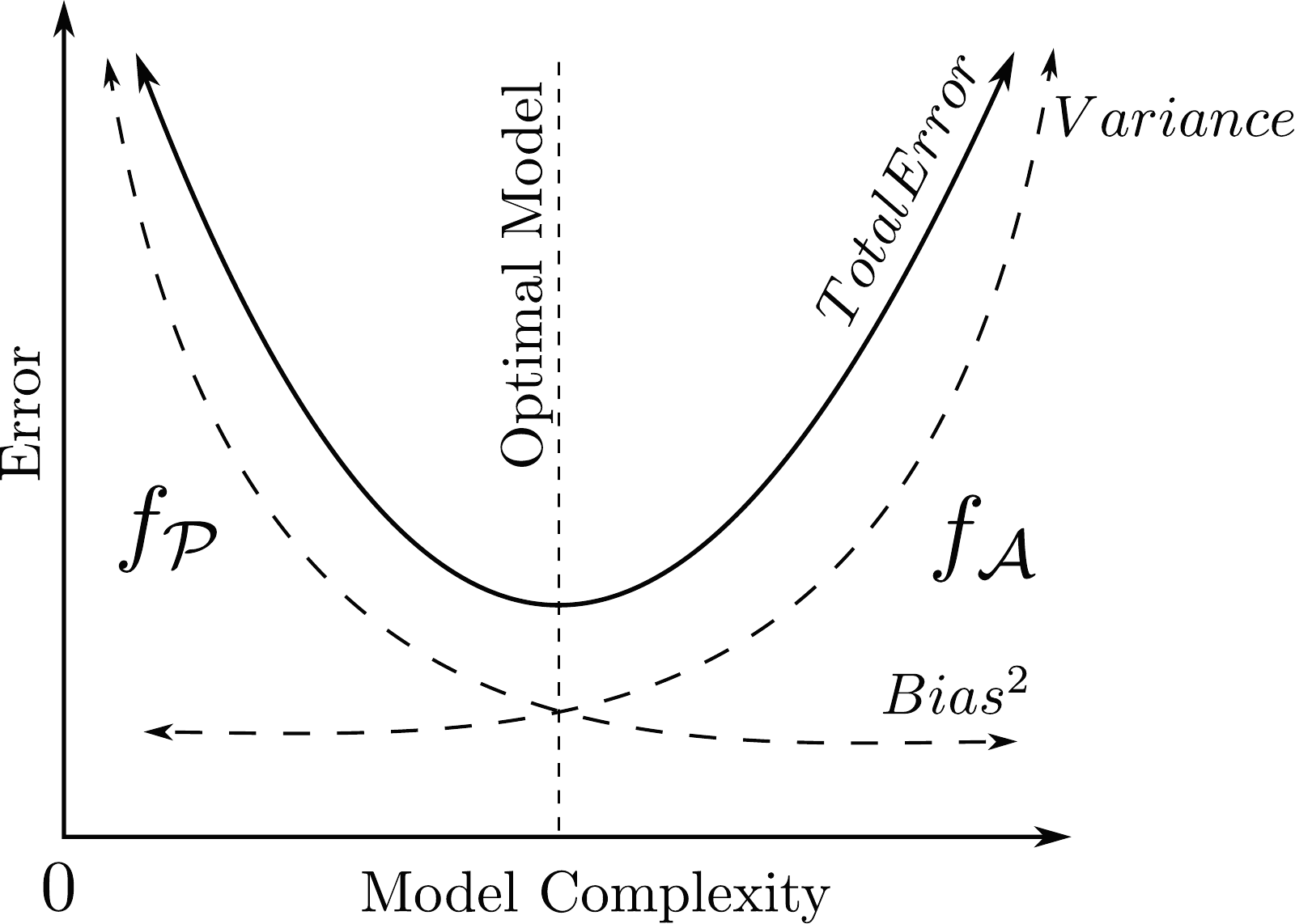}
\caption{ Model complexity vs. Total Error. The optimal models are likely to belong to the class of PEAI models. }
\label{Fig:BiasVariance}
\end{figure}

Normally, a physics models tend to have higher bias, but low variance. On the other hand, AI models tend to have a high variance and low bias. By placing physics constraints on the AI model during learning or during run-time, one places a bias on the new hybrid model, limiting the output space. This will cause the model to have a larger bias, and if done correctly will dramatically reduce the variance of the hybrid model.This concept can be shown graphically by model complexity vs. error diagram in Figure \ref{Fig:BiasVariance}. By intelligently introducing physics based constraints to AI models or vice versa, we can arrive at models that have lower total error.

\section{ Implications of PEAI } 

\begin{figure}[ht!]
\centering
\includegraphics[width=0.7\textwidth]{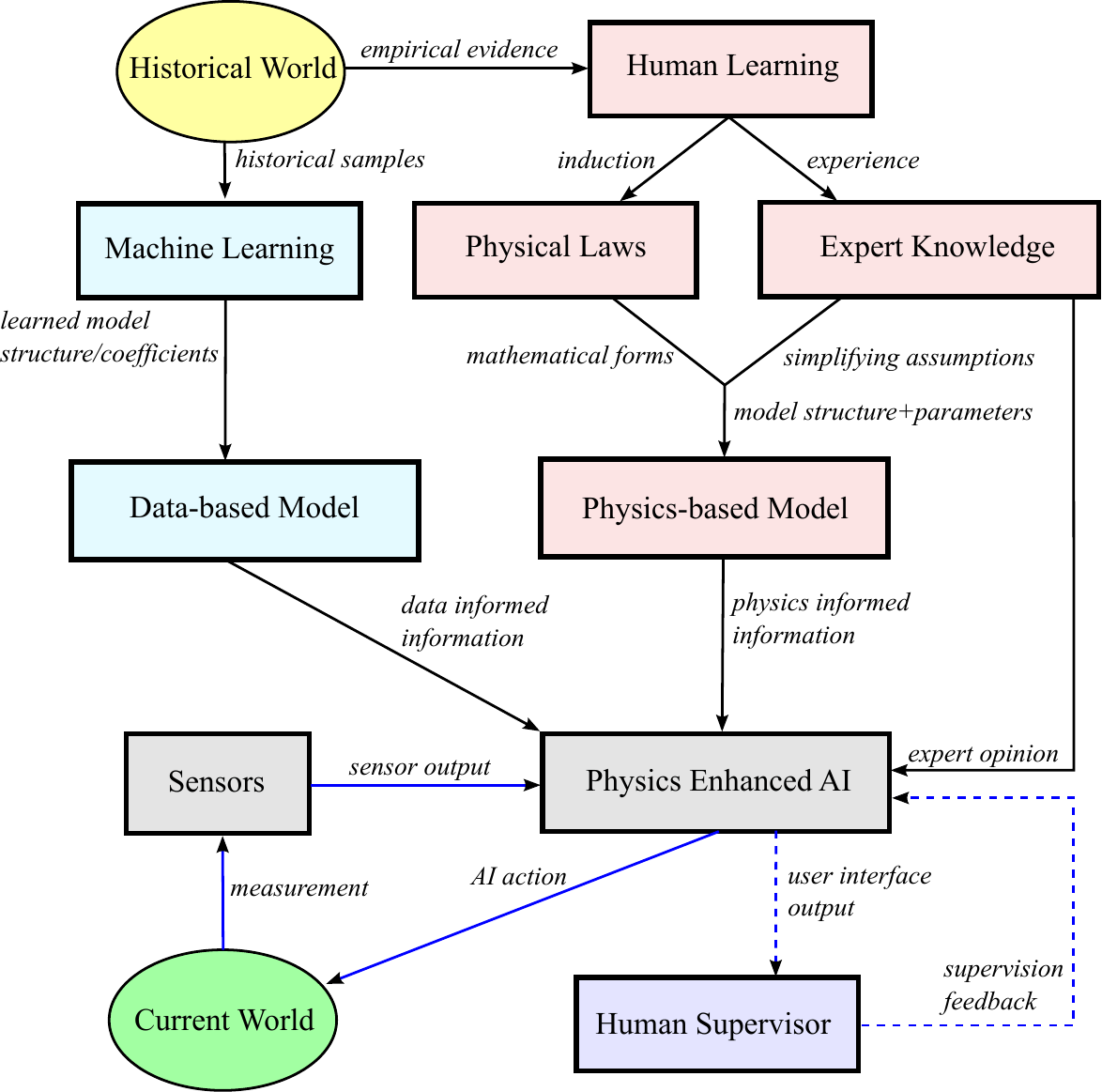}
\caption{ Physics enhanced AI and its relationships with machine learning and human learning }
\label{Fig:Arch}
\end{figure}

Human learning builds on human observations and empirical evidence of the surrounding world, and this accumulated and learned knowledge is passed on, resulting in a systems design or model that is physics-based, as shown by the architecture in Figure \ref{Fig:Arch}. Similarly, data driven AI approaches use ML to arrive at a design or model based on the collected data. The combined model allows solutions of AI to be constrained by physical solution and expert knowledge, which enhances performance and trust. In addition, user trust in PEAI can be further enhanced by having a human supervisor in the loop, where the supervisor monitors the AI and provides feedback to the system that can improve its performance.

It is interesting to point out that many have already worked on the class of PEAI systems, though the direct increase in user trust was not the motivation. For example, AI system with physical constrains is shown best by the work  \textit{Physics informed deep learning} \cite{Raissi:2017a,Raissi:2017b}, where differential equation constraints that can represent a physical model is combined with a neural network to form a PEAI. In the field of predictive turbulence modeling, Physics-informed ML framework has been proposed \cite{wang2017comprehensive}, where the functional form of the Reynolds stress discrepancy in Reynolds-averaged Navier-Stokes (RANS) is learned directly based on available data. In the area of real-time vision-based event monitoring at industrial sites, Physical constraints are added to AI models in order to improve performance and enhance user trust. 

Being able to quickly identify and design AI solutions that have potential for wide-range industrial adoption is challenging. The AI solutions that are more likely to receive wider adoption are ones that can earn trust from users, and deliver an improved productivity at the same time. By explicitly pointing out the connection between enhanced user trust and combining models from different domains, it is suggested that one should always seek to combine AI models with prior models, if available. 

\section{ Conclusions }

Physics enhanced AI (PEAI) is a class of model that is formed by intelligently combining models from the domains of artificial intelligence or machine learning with physical and expert models. It was shown that by doing so, model risk is reduced, resulting in a more ``trustworthy'' model than any one model from a single domain. PEAI is shown as a solution to improve user adoption of an AI which solves a complex yet narrow enough problem.


\bibliographystyle{ieeetr} 
\bibliography{ref} 
\end{document}